\documentclass[letterpaper, 10 pt, conference]{ieeeconf}  

\IEEEoverridecommandlockouts                              

\usepackage{multirow}
\usepackage{makecell}
\usepackage{amsmath}
\usepackage{amssymb}
\usepackage{bm}
\usepackage{graphicx}
\usepackage{xcolor}
\usepackage{booktabs}  
\usepackage{multirow}  
\usepackage{makecell}   
\usepackage{siunitx}    
\usepackage{subcaption}
\usepackage{hyperref}
\newcommand\conf{\bm{q}}
\newcommand\state{\bm{x}}
\newcommand\control{\bm{u}}
\newcommand\velocity{\bm{v}}
\newcommand\accel{\bm{a}}

\title{\LARGE \bf
COSMIK-MPPI: Scaling Constrained Model Predictive Control to Collision Avoidance in Close-Proximity Dynamic Human Environments
}

\author{Ege Gursoy$^{1*}$,  Maxime Sabbah$^{1*}$, Arthur Haffemayer$^{2}$, Joao Cavalcanti Santos$^{3}$, \\ Pietro Noah Crestaz$^{1,6}$, Vladimir Petrik$^{2}$, Nicolas Mansard$^{1,4}$, Vincent Bonnet$^{1,5}$
\thanks{* equal contribution, corresponding author: \texttt{egursoy@laas.fr}}
\thanks{$^{1}$ LAAS-CNRS, Université de Toulouse, CNRS, Toulouse, France}%
\thanks{$^{2}$ CIIRC, Czech Technical University, Prague}%
\thanks{$^{3}$ LIRMM, Université de Montpellier, CNRS, Montpellier, France}%
\thanks{$^{4}$ Artificial and Natural Intelligence Toulouse Institute (ANITI), Toulouse, France}%
\thanks{$^{5}$ Image and Pervasive Access Laboratory (IPAL), CNRS-UMI, 2955,
Singapore}%
\thanks{$^{6}$ Industrial Engineering Department, University of Trento, Trento, Italy}
}

\begin{document}

\maketitle
\thispagestyle{empty}
\pagestyle{empty}

\begin{abstract}

Ensuring safe physical interaction between torque-controlled manipulators and humans is essential for deploying robots in everyday environments. Model Predictive Control (MPC) has emerged as a suitable framework thanks to its capacity to handle hard constraints, provide strong guarantees and zero-shot adaptability through predictive reasoning. However, Gradient-Based MPC (GB-MPC) solvers have demonstrated limited performance for collision avoidance in complex environments. 
Sampling-based approaches such as Model Predictive Path Integral (MPPI) control offer an alternative via stochastic rollouts, but enforcing safety via additive penalties is inherently fragile, as it provides no formal constraint satisfaction guarantees.
We propose a collision avoidance framework called COSMIK-MPPI combining MPPI with the the toolbox for human motion estimation RT-COSMIK and the Constraints-as-Terminations transcription, which enforces safety by treating constraint violations as terminal events, without relying on large penalty terms or explicit human motion prediction. 
The proposed approach is evaluated against state-of-the-art GB-MPC and vanilla MPPI in simulation and on a real manipulator arm. Results show that COSMIK-MPPI achieves a 100\% task success rate with a constant computation time (22 ms) largely outperforming GB-MPC. 
In simulated infeasible scenarios, COSMIK-MPPI consistently generates collision-free trajectories, contrary to vanilla MPPI. 
These properties enabled safe execution of complex real-world human–robot interaction tasks in shared workspaces using an affordable markerless human motion estimator, demonstrating a robust, compliant and practical solution for predictive collision avoidance (cf. results showcased in \url{https://exquisite-parfait-ffa925.netlify.app}).
\end{abstract}

\section{Introduction}

Trajectory planning and safe control of robotic manipulators in human-shared environments remains challenging due to nonlinear dynamics, sensing uncertainty, and strict safety requirements. In human–robot interaction, robots must guarantee collision avoidance under limited reaction time and noisy perception, while operating in dynamic environments shaped by hard-to-predict human motion. These constraints make anticipatory and robust motion generation a central challenge for collaborative manipulation \cite{robla2017working,li2024safe}.

Collision avoidance for articulated manipulators has been widely studied using reactive and predictive approaches \cite{de2012integrated, 8665083, Safeea2019Online,  Oelerich2026SafeFlowMPC}. Reactive methods, such as damped least-squares velocity filters or potential-field repulsion strategies, are computationally efficient but remain fundamentally myopic. As they do not explicitly reason over a planning horizon, they can exhibit oscillations, conservative behaviors near narrow passages, or unstable avoidance in dynamic scenarios \cite{de2012integrated, Safeea2019Online}. Predictive control methods, on the other hand, enable horizon-based anticipation and principled trade-offs between task execution, dynamics, and safety \cite{11128856, Oelerich2026SafeFlowMPC}. In torque-controlled systems, such formulations further allow collision avoidance to be handled directly at the level of actuator constraints. Most predictive collision avoidance approaches rely on Gradient-Based Model Predictive Control (GB-MPC) formulations, which have mainly relied on second order optimization solvers \cite{mastalli20crocoddyl, verschueren2022acados}, yet raising several bottlenecks. First of all, the non-convexity of collision constraints typically induce multiple homotopy classes of feasible solutions. As a result, GB-MPC solvers are highly sensitive to initialization, may converge to suboptimal local minima, or may fail to meet real-time constraints due to the associated computational burden. These issues were highlighted in the works introducing diffusion-warm-start MPC \cite{Haffemayer2026DiffusionMPC} and SafeFlowMPC \cite{Oelerich2026SafeFlowMPC}. Secondly, GB-MPC poorly scales when facing many constraints simultaneously saturating. As collision pairs must be created between several convex shapes, a typical robotic application deals with multiple collision constraints, for which GB-MPC has a hard time to handle, as it will be shown in the present study. 

\begin{figure}[t]
  \centering
    \includegraphics[
      width=\linewidth,
      trim=0 0.62cm 0 6.9cm,
      clip
  ]{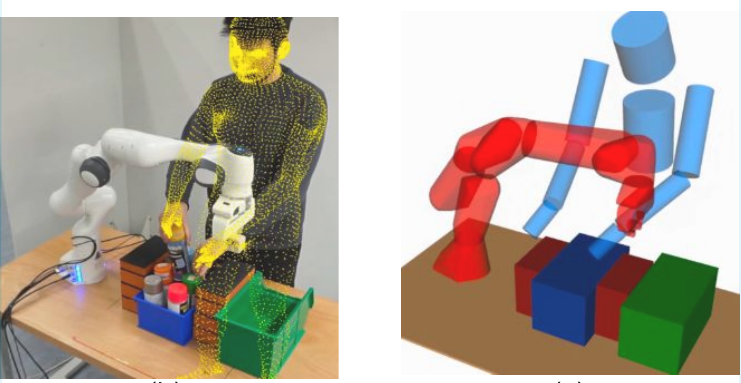}
\caption{Visualization of the NLF-based human mesh recovery \cite{sarandi2024nlf} (left) and 
human collision capsules (rendered as cylinders only visually) and output of the inverse kinematics pipeline used to position collision primitives (right).}
  \label{fig:overview}
  
  \vspace{-0.6cm}
\end{figure}

An additional limitation is the reliance on accurate obstacle motion estimation and prediction. Optimization-based predictive controllers typically assume access to present and future obstacle trajectories, which is unrealistic in collaborative scenarios characterized by noisy sensing, partial observability, and non-stationary human behavior. This issue is further exacerbated by markerless human pose estimation. Despite recent progress, including volumetric human shape representations such as Neural Localizer Fields (NLF) \cite{sarandi2024nlf}, markerless approaches remain sensitive to occlusions, viewpoint changes, and latency, leading to noisy and biomechanically inconsistent pose estimates. These errors propagate directly to collision constraints, forcing a trade-off between overly conservative safety margins and optimistic predictions that may compromise safety.

Sampling-based predictive control methods, such as Model Predictive Path Integral (MPPI) control, offer an alternative by evaluating large sets of control perturbations through forward rollouts of the system dynamics, rather than relying on local gradient information \cite{7487277}. This enables MPPI to naturally handle non-smooth objective functions and explore multiple candidate futures in parallel, reducing sensitivity to local minima induced by collision constraints \cite{williamsRobustSamplingBased2018}. Moreover, uncertainty in human pose estimation can be incorporated directly by sampling obstacle configurations from noisy measurements, without committing to a single predicted human trajectory. Recent works have demonstrated the feasibility of MPPI for high-dimensional manipulators in rather simple scenarios involving dynamic obstacles \cite{Zhou2025ParallelMPPI}.

However, in most MPPI formulations, safety constraints are handled through additive penalty terms in the running costs. Finite penalties cannot reliably prevent constraint violations, while large penalties introduce severe tuning difficulties and numerical instability. Furthermore, early termination of violating rollouts often used to reduce computation, can artificially reduce accumulated cost, inadvertently encouraging unsafe behaviors. To address these limitations, Constraints as Terminations (CaT) \cite{chane2024cat} was proposed as a principled mechanism that treats constraint violations as  terminal events, making unsafe trajectories highly unlikely under the sampling distribution. While CaT was originally introduced in reinforcement learning, constraint discounting within MPPI was first proposed in TD-CD-MPPI \cite{crestaz2025td}, where it was combined with value-function learning for locomotion control.

The present work uses CaT by proposing a compliant COSMIK-MPPI formulation for predictive collision avoidance for human–robot interactions in shared workspace. The method operates with noisy, markerless human perception and avoids explicit human motion prediction. In light of the above limitations of existing predictive collision avoidance methods for robotic manipulators, the contributions of this paper are:
\begin{itemize}
    \item A new COSMIK-MPPI controller for complex dynamic collision avoidance. 
    \item A systematic comparison between state-of-the-art GB-MPC \cite{haffemayer2024model}, vanilla MPPI \cite{williams2017model}, and the proposed COSMIK-MPPI formulation for collision avoidance, analyzing robustness to local minima, sensitivity to human motion uncertainty, compliance and real-time feasibility in simulation and on a real robotic manipulator arm.
    \item A demonstrator for human–robot interactions in shared workspace that operates directly with markerless, noisy human pose estimates.
\end{itemize}

\section{Methods}
 
\subsection{Gradient-Based Model Predictive Control}

The Optimal Control Problem (OCP) solved in this part, is identical to the one described in \cite{Haffemayer2026DiffusionMPC}:
\begin{subequations}\label{equ:ocp}
\label{sqp_equation}
\begin{align}
 \underset{\state , \control }{\min} & \sum^{T-1}_{t = 0} \ell_t (\state_t, \control_t) + \ell_T (\state_T)\\
\text{subject to   }
    \state_{t+1} &= f_t(\state_t,\control_t)~~\forall~ 0 \leq t < T \\
    c_t(\state_t,\control_t) &\geq 0,~~\forall~ 0 \leq t < T\\
    c_T(\state_T) &\geq 0
\end{align}
\end{subequations}

\noindent where $ \state = (\state_0, \dots, \state_T)$ and $\control = (\control_0, \dots, \control_{T-1})$ are the discretized state-control sequences, and more specifically, $\state_t = (\conf_t, \velocity_t)$ is the robot state and $\control_t$ is the joint torques at time $t$; $f_t$ is the discretized robot dynamics model along the horizon of length $T$; $\ell_t$ are the running costs and $\ell_T$ is the terminal cost; $c_t$ and $c_T$ represent the inequality constraints (comprising the collision avoidance ones defined as in \cite{haffemayer2024model}) for the running and the terminal nodes. 

The OCP running costs $\ell_t$ for the time $t$ are composed of:

\begin{itemize}
\item Goal reaching task cost: 
    \begin{equation}\label{equ:goal_reaching}
    \ell_{ee}(\state_t) = || \log(T_{\text{goal}}^{-1} \cdot T_{\text{ee}}(\conf_t))  ||_{Q_{ee}}^2 
\end{equation}
    \item State regularization cost: 
\begin{equation}\label{equ:state_regu}
    \ell_{\state}(\state_t) = ||\state_t - \state_0||_{Q_{\state}}^2  
\end{equation}
    \item Control regularization cost:
    \begin{equation}\label{equ:control_regu}
    \ell_{\control}(\state_t, \control_t) = ||\control - \control_\text{{grav}}(\conf_t)||_{Q_{\control}}^2 
\end{equation}
\end{itemize}

\noindent with $Q_{ee}\boldsymbol{\succeq}0$ a weight matrix setting the relative importance of translational and rotational errors; $Q_{\state}\boldsymbol{\succeq}0$ and $Q_{\control}\boldsymbol{\succeq}0$ the weight diagonal matrices of the state and control penalizations; $T_{\text{goal}} \in \textit{SE(3)}$ is the pose of the target; $\control_\text{{grav}}$ is the gravity compensation torque computed from the robot dynamics model.
The terminal cost $\ell_T$ is defined the same without the control regularization. The OCP is solved in a receding-horizon scheme and implemented within the Crocoddyl library \cite{mastalli20crocoddyl} using the constrained Sequential Quadratic Programming (SQP) solver from the mim\_solver library \cite{jordana2025structure}.

\subsection{Vanilla Model Predictive Path Integral}

As a sampling-based baseline, we consider the STORM \cite{bhardwaj2022storm} formulation of MPPI, which performs receding-horizon optimization in joint space. At each control cycle, the controller samples a batch of open-loop control sequences, rolls them out through an approximate model of the manipulator, evaluates their finite-horizon costs, and updates the sampling distribution toward lower-cost trajectories.

Reusing the formulation introduced in the previous section, MPPI maintains a Gaussian distribution over control sequences :
\begin{equation}
\pi_{\bm{\mu}}=\prod_{t=0}^{T-1}
\mathcal{N}(\control_t;\bm{\mu}_t,\bm{\Sigma}_t) 
\label{eq:mppi_policy}
\end{equation}
parameterized by the sequence of means
$\bm{\mu} = (\bm{\mu}_0,\dots,\bm{\mu}_{T-1})$
and covariance matrices
$\bm{\Sigma} = (\bm{\Sigma}_0,\dots,\bm{\Sigma}_{T-1})$.

At each control update, $K$ control sequences are sampled from \eqref{eq:mppi_policy} and propagated through the approximate dynamics model. Starting from the measured initial state $\state_0$, the rollout is formulated as:
\begin{align}
    \control^k &= (\control_{0,k}, \control_{1,k}, \dots, \control_{T-1,k}), && k=1,\dots,K \label{eq:mppi_samples} \\
    \state_{t+1,k} &= f_t(\state_{t,k}, \control_{t,k}), && t=0,\dots,T-1  \label{eq:mppi_rollout}
\end{align}

Each rollout is assigned a discounted finite-horizon cost:
\begin{equation}
L_k=
\sum_{t=0}^{T-1}
\gamma_t \, \ell_{t,k}
+
\gamma_{T}\ell_{T,k}  
\label{eq:mppi_rollout_cost}
\end{equation}
where $\gamma \in [0,1]$ is a discount factor, for rollout $k$, $\ell_{t,k}$ is the running cost at timestep $t$, and $\ell_{T,k}$ is the terminal cost. In this vanilla formulation,
\begin{equation}
\ell_{t,k}
=
\ell_{ee}(\state_{t,k})
+
\ell_{\state}(\state_{t,k})
+
\ell_{\control}(\state_{t,k},\control_{t,k})
+
\ell_{\mathrm{coll}}(\state_{t,k})
\label{eq:mppi_stage_cost}
\end{equation}
with $\ell_{ee}$, $\ell_{\state}$, and $\ell_{\control}$ the costs respectively introduced in \eqref{equ:goal_reaching}, \eqref{equ:state_regu}, and \eqref{equ:control_regu}. The additional collision penalty cost is computed from convex approximations of the robot links, the environment, and the human body. Let $d^{\mathrm{sgn}}_{r}(\state_{t,k})$ denote the signed distance associated with collision pair $r$, where negative values indicate free space and positive values indicate overlap. To ensure a non-negative collision cost, the cost is defined as: 
\begin{equation}
\ell_{\mathrm{coll}}(\state_{t,k})
=
\sum_{r}
\max\!\left(
d^{\mathrm{sgn}}_{r}(\state_{t,k}) + d_{\mathrm{th}},
\,0
\right)
\label{eq:mppi_collision}
\end{equation}
where $d_{\mathrm{th}}>0$ is the safety-margin threshold.

Following MPPI, the rollout weights are computed as
\begin{equation}
\eta_k
=
\frac{
\exp\!\left(
-\frac{1}{\beta}
\left(
L_k - \min_{i=1,\dots,K} L_i
\right)
\right)
}{
\sum_{j=1}^{K}
\exp\!\left(
-\frac{1}{\beta}
\left(
L_j - \min_{i=1,\dots,K} L_i
\right)
\right)
}
\label{eq:mppi_weights}
\end{equation}
where $\beta > 0$ is the temperature parameter. These weights define a soft-min over sampled trajectories and assign larger importance to lower-cost rollouts.

Finally, the control sequence is updated as a weighted average of the sampled controls
\begin{equation}
\bm{\mu}_h
\leftarrow
(1-\alpha_{\mu})\,\bm{\mu}_h
+
\alpha_{\mu}
\sum_{k=1}^{K}\eta_k \control_{t,k},
~~ t=0,\dots,T-1 
\label{eq:mppi_mean_update}
\end{equation}
where $\alpha_{\mu}\in(0,1]$ is a step size.

The control is applied in a receding-horizon manner. Only the first control of the optimized sequence is applied, after which the horizon is shifted and the optimization is repeated from the new measured state. This allows MPPI to react online to perception noise, model mismatch, and changes in the human configuration.

\subsection{COSMIK--Model Predictive Path Integral}

To replace the additive collision penalty used in vanilla MPPI, the proposed formulation relies on CaT \cite{chane2024cat}. The motivation is that, in a standard additive-cost formulation, collision avoidance remains a soft trade-off against the task objective and therefore depends strongly on the tuning of the collision weight. In contrast, CaT treats collision proximity as a progressive loss of future utility. Rollouts that come too close to the obstacle become less competitive over the remainder of the horizon. The task-related stage and terminal costs are therefore defined separately as
\begin{align}
    \ell^{\mathrm{task}}_{t,k} &=
    \ell_{ee}(\state_{t,k}) + \ell_{\state}(\state_{t,k}) +
    \ell_{\control}(\state_{t,k},\control_{t,k})
    \label{eq:cat_task_cost} \\
    \ell^{\mathrm{task}}_{T,k} &=
    \ell_{ee}(\state_{T,k}) + \ell_{\state}(\state_{T,k})
    \label{eq:cat_terminal_task_cost}
\end{align}

The collision constraint provides a dense nonnegative violation magnitude
\begin{equation}
v^c_{t,k}
=
\ell_{\mathrm{coll}}(\state_{t,k}),
\label{eq:cat_violation}
\end{equation}
with $\ell_{\mathrm{coll}}$ defined in \eqref{eq:mppi_collision}. Intuitively,
$v^c_{t,k}=0$ corresponds to a safe rollout at stage $t$, while increasing
values indicate that the rollout is approaching or violating the collision
margin. This violation is mapped to a per-stage termination hazard
\begin{equation}
\delta^c_{t,k}
=
p^c_{\max}
\,\mathrm{clip}\!\left(
\frac{v^c_{t,k}}{c^c_{\max}},\,0,\,1
\right)
\label{eq:cat_delta}
\end{equation}
where $p^c_{\max}\in(0,1]$ controls the maximum termination probability and
$c^c_{\max}$ is updated online as an exponential moving average of the maximum
observed violation:
\begin{equation}
c^c_{\max}
\leftarrow
\tau^c_c\, c^c_{\max}
+
(1-\tau^c_c)\max_{t,k} v^c_{t,k}
~~ \tau^c_c \in [0,1)
\label{eq:cat_cmax}
\end{equation}
This normalization makes the hazard adaptive to the scale of the currently
observed violations and avoids having to hand-tune a fixed normalization
constant.

The corresponding rollout survival factor is then
\begin{equation}
S_{t,k}
=
\prod_{m=0}^{t}(1-\delta^c_{m,k})
\label{eq:cat_survival}
\end{equation}
which can be interpreted as the probability that rollout $k$ has remained
collision-safe up to stage $t$. As a result, trajectories that repeatedly
approach collision lose future influence in the MPPI weighting, whereas safe
trajectories preserve their contribution over the full horizon. This provides a
stronger safety bias than a standard additive collision term while retaining a
dense signal for optimization.

A direct multiplication of the MPPI costs by $S_{t,k}$ would be incorrect in a cost-minimization setting, since terminating a rollout early would also stop the accumulation of future costs. To preserve the CaT interpretation, we introduce a shifted positive reward associated with the task cost:
\begin{equation}
\begin{aligned}
r_{t,k} &= b - \ell^{\mathrm{task}}_{t,k} \\
r_{T,k} &= b_T - \ell^{\mathrm{task}}_{T,k}
\end{aligned}
\label{eq:cat_shifted_rewards}
\end{equation}
where $b$ and $b_T$ are baselines chosen such that
$r_{t,k}\ge 0$ and $r_{T,k}\ge 0$ for all sampled rollouts. In practice, the
baselines are updated online using upper-envelope exponential moving averages:
\begin{equation}
\begin{aligned}
\bar{b}
&=
\max_{t,k}\ell^{\mathrm{task}}_{t,k},
&
b
&\leftarrow
\max\!\big(\tau_b b + (1-\tau_b)\bar{b},\, \bar{b}\big) + \varepsilon
\\
\bar{b}_T
&=
\max_{k}\ell^{\mathrm{task}}_{T,k},
&
b_T
&\leftarrow
\max\!\big(\tau_b b_T + (1-\tau_b)\bar{b}_T,\, \bar{b}_T\big) + \varepsilon
\end{aligned}
\label{eq:cat_b}
\end{equation}
where $\tau_b \in [0,1)$ and $\varepsilon>0$ is a small numerical constant.

The modified trajectory score is then written as the pseudo-cost
\begin{equation}
L_k^{\mathrm{CaT}}
=
-
\sum_{t=0}^{T-1}\gamma_t S_{t,k} r_{t,k}
-
\gamma_{T} S_{T,k} r_{T,k}
\label{eq:cat_score}
\end{equation}
Equivalently,
$-S_{t,k}r_{t,k}$ and $-S_{T,k}r_{T,k}$ can be interpreted as running and terminal pseudo-costs, which allows the MPPI cost-to-go to be reused.

The importance weight of rollout $k$ is finally computed as
\begin{equation}
\eta_k
=
\frac{
\exp\!\left(
-\frac{1}{\beta}
\left(
L_k^{\mathrm{CaT}}
-
\min_{i=1,\dots,K} L_i^{\mathrm{CaT}}
\right)
\right)
}{
\sum_{j=1}^{K}
\exp\!\left(
-\frac{1}{\beta}
\left(
L_j^{\mathrm{CaT}}
-
\min_{i=1,\dots,K} L_i^{\mathrm{CaT}}
\right)
\right)
}
\label{eq:cat_weights}
\end{equation}

This formulation preserves the sampling-based nature of MPPI and replaces additive collision penalties by terminations. Rollouts that approach the human too closely experience a drop in survival and therefore lose future contribution, which strongly reduces their importance weights without requiring excessively large collision costs. For the whole study, we call this framework COSMIK-MPPI to differentiate it from vanilla MPPI.

\section{Implementation details}

\subsection{Real-time human motion estimation}

\begin{figure}[t]
  \centering
    \includegraphics[
      width=0.8\linewidth,
      trim=0 7.7cm 0 0,
      clip
  ]{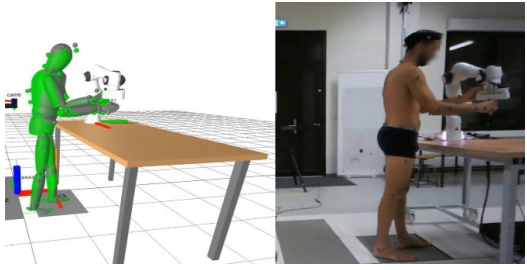}
\caption{Experimental validation, reported in \cite{rtcosmik2025}, comparing RT-COSMIK human motion estimation (green) with a reference marker-based optical motion capture system (grey) during a collaborative human–robot scenario.}
  \label{fig:cosmik}
  
  \vspace{-0.5cm}
\end{figure}

Human motion estimation was performed using RT-COSMIK \cite{rtcosmik2025}, a markerless solution based on two RGB cameras and a biomechanically constrained inverse kinematics pipeline. Across a range of industrial tasks, joint angles are estimated with an average root-mean-square error of approximately 7 degrees. RT-COSMIK is designed to be robust to noisy and incomplete observations, making it well suited for human–robot interaction scenarios. 

In its latest version, RT-COSMIK is based on NLF \cite{sarandi2024nlf}, which represent the human body as a continuous implicit field mapping 3D spatial locations to body-part occupancy, enabling pose and shape estimation without explicit keypoint detection. By formulating localization as dense field regression, NLF avoids brittle keypoint associations and remains robust to occlusions and self-collisions. Inference relies on efficient neural field queries followed by lightweight optimization, enabling real-time operation (10 to 30 ms). Fig.\ref{fig:cosmik} illustrates representative accuracy and robustness results, extracted from \cite{rtcosmik2025}, comparing RT-COSMIK markerless estimates against a reference marker-based optical motion capture system. 

RT-COSMIK pipeline is based on 3 steps:
\begin{enumerate}
    \item human pose estimation: relying on GPU TensorRT YOLO model \cite{yolov10} for human detection and on NLF model for estimating a dense set of 2D body keypoint positions expressed in each camera frame (Fig.~\ref{fig:overview}). 
    \item triangulation and filtering: which converted the 2D keypoints extracted in multiple cameras into 3D keypoints using a direct linear transform with the calibrated camera parameters;
    \item inverse kinematics: which uses a full body biomechanical model to recover the vector of joint angles matching at best the virtual markers solving a constrained quadratic programming problem.
\end{enumerate}

The overall frequency of the pipeline was benchmarked at 20Hz with two rolling shutter RGB webcams (1280 x 720 MJPEG, 40fps) positioned approximatly at 3 m of the scene to see the participant's upper limbs. 

\subsection{Collision Avoidance}

For real-time performance, both the robot and the human body are approximated as unions of strictly convex primitives, primarily capsules.  Capsules provide a tight and smooth approximation of articulated limbs while preserving closed-form distance computation between pairs. The signed distance between two capsules admits an analytical solution based on segment-to-segment distance, thus enabling high-frequency evaluation during rollouts. As shown in Fig. \ref{fig:overview}, 11 capsules are attached to the robot (red), 5 to the scene (table and boxes), and 8 to the human (blue, rendered as cylinders for visuals only). In that case, the control scheme consider a total of 24 collisions pairs.  

\subsection{Low-level controller}
\label{subsec:low_level_controller}
For fairness, the comparison between GB-MPC and MPPI has been performed using the low-level joint torque controller available on the robot. Yet, the torques references and the gains used are differently obtained. 

\paragraph{GB-MPC}
The dynamic model $f(\state,\control)$ in Eq.~\ref{sqp_equation}.b is implemented as a feed-forward torque integration, where $\control$ is the joint torque, following the classical approach of GB-MPC litterature \cite{mastalli20crocoddyl, sleiman2021unified}. As the GB-MPC steps take longer than 1ms (the direct low-level control frequency), the Ricatti gains are used to interpolate between two solver iterations. The commanded joint torques are then:
\begin{equation}
    \boldsymbol{\tau} = \boldsymbol{u}^\star(\state_0)
    + \mathbf{K}^{\star}(\state_0)\left(\state - \state_0\right)
    \label{eq:low_level_mpc}
\end{equation}
\noindent where $\state$ is the latest state measurement, $\state_0$ is the state measurement available when the latest OCP solution has been computed; $\control^\star$, $\mathbf{K}^\star$ are given by the OCP and its derivatives, and called the feedforward and the feedback Ricatti gains. The resulting controller has the shape of an impedance controller, whose gains are defined by the resolution of the problem. 

\paragraph{MPPI}
Following the MPPI literature \cite{williamsRobustSamplingBased2018, Zhou2025ParallelMPPI}, the rollout dynamics model $f_t$ in \eqref{eq:mppi_rollout} is implemented as a joint-space second-order integrator with acceleration control. The control input $\control_t=\accel_t$ is interpreted as a desired joint acceleration. MPPI returns an updated horizon-length control sequence $\{\control^\star_{t+h}\}_{h=0}^{T-1}$, of which only the first element, denoted $\control^\star(\state_0)$, is applied. This command is integrated online with step $\Delta t$ to produce the next reference state $\state_1^\star(\state_0,\control^\star(\state_0))$. The commanded joint torques between two solver iterations are then:
\begin{equation}
\boldsymbol{\tau}
=
\boldsymbol{\tau}^{\mathrm{ff}}
+
\mathbf{K}_{\mathrm{fix}}\!\left(
\state_1^\star\!\left(\state_0,\control^\star(\state_0)\right) - \state
\right)
\label{eq:low_level_lf}
\end{equation}
\noindent where $\boldsymbol{\tau}^{\mathrm{ff}}=\mathrm{RNEA}\!\left(\conf_1^\star,\velocity_1^\star,\control^\star(\state_0)\right)$ is the feedforward inverse-dynamics torque typically obtained thanks to the Pinocchio library \cite{carpentier2019pinocchio}, and $\mathbf{K}_{\mathrm{fix}}$ is a fixed state-feedback gain acting on the tracking error.

\section{Simulation benchmark}
 \begin{figure}[t]
  \centering
  \includegraphics[width=0.6\linewidth]{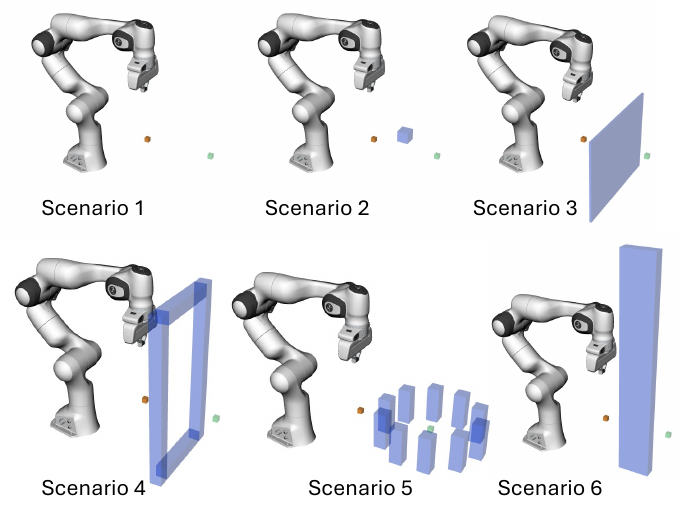}
 \caption{Views of the 6 scenarios used to assess collision avoidance abilities in simulation.The red and green dots are the alternative targets.}
  \label{fig:simulation}

  \vspace{-0.5cm}

\end{figure}
As shown in Fig.~\ref{fig:simulation}, predictive collision avoidance is evaluated in simulation using six scenarios of increasing geometric complexity, all based on the same pick-and-place task and sharing identical task parameters and initial warm starts for GB-MPC and MPPI:

\begin{itemize}
\item Scenario 1: Baseline pick-and-place task 
\item Scenario 2: A single static collision capsule is introduced in the workspace.
\item Scenario 3: The collision capsule is elongated, increasing geometric complexity and restricting feasible avoidance homotopies.
\item Scenario 4: Four collision primitives create a non-convex configuration.
\item Scenario 5: A dense circular arrangement of collision capsules emulates a highly constrained, strongly non-convex environment and stresses computational scalability.
\item Scenario 6: A long collision box blocks the direct path to the target, rendering the task geometrically infeasible and serving as a negative-control benchmark.
\end{itemize}

Performance is evaluated using the following metrics:
\begin{itemize}
\item Task success rate: Fraction of trials completed without collision \cite{Haffemayer2026DiffusionMPC,Oelerich2026SafeFlowMPC, Gafur2021Dynamic}.
\item Distance to target: Final Euclidean distance between the end-effector and the target pose.
\item Distance to obstacles: Minimum signed distance to collision objects along the trajectory, averaged across trials \cite{Zhou2025ParallelMPPI}.
\item Trajectory smoothness: Integrated squared joint torque variation.
\item Computation time: Average control computation time per planning step \cite{williams2017model}.
\end{itemize}

\begin{table}[t]
\centering
\setlength{\tabcolsep}{1.3pt}
\caption{Simulation comparison of GB-MPC, vanilla MPPI and COSMIK-MPPI for collision avoidance.}
\label{tab:predictive_collision_comparison}
\begin{tabular}{@{} l l c S[table-format=0.0] S[table-format=0.0] S[table-format=0.0] S[table-format=0.0] @{}}
\toprule
\textbf{Sc.} & \textbf{Method} & \makecell{\textbf{Success}\\\textbf{Rate}} & {\makecell{\textbf{Dist. to}\\\textbf{Target (cm)}}} & {\makecell{\textbf{Dist. to}\\\textbf{Obst. (cm)}}} & {\makecell{\textbf{Smooth-}\\\textbf{ness}}} & {\makecell{\textbf{Comp.}\\\textbf{Time (ms)}}} \\ 
\midrule

\multirow{3}{*}{1} 
 & GB-MPC      & 1.00 & 1.05  & {--}  & 1.50  & 15   \\
 & Vanilla MPPI     & 1.00 & 1.88  & {--}  & 18.59 & 22   \\
 & COSMIK-MPPI & 1.00 & 1.98  & {--}  & 20.14 & 22   \\ \addlinespace

\multirow{3}{*}{2} 
 & GB-MPC      & 1.00 & 0.51  & 8.78  & 2.31  & 89   \\
 & Vanilla MPPI     & 1.00 & 1.79  & 9.32  & 20.83 & 22   \\
 & COSMIK-MPPI & 1.00 & 1.62  & 10.65  & 17.54 & 22   \\ \addlinespace

\multirow{3}{*}{3} 
 & GB-MPC      & 1.00 & 0.26  & 4.69 & 2.18  & 81   \\
 & Vanilla MPPI     & 1.00 & 1.95  & 4.09 & 24.65 & 22   \\
 & COSMIK-MPPI & 1.00 & 1.83  & 5.35  & 19.97 & 22   \\ \addlinespace

\multirow{3}{*}{4} 
 & GB-MPC      & 0.00 & 54.21 & 23.10 & 0.00  & 415  \\
 & Vanilla MPPI     & 1.00 & 1.91  & 3.57 & 24.57 & 22   \\
 & COSMIK-MPPI & 1.00 & 1.95  & 1.62 & 20.72 & 22   \\ \addlinespace

\multirow{3}{*}{5} 
 & GB-MPC      & 0.00 & 73.97 & 14.87  & 0.00  & 1812 \\
 & Vanilla MPPI     & 1.00 & 1.86  & 4.35 & 17.82 & 22   \\
 & COSMIK-MPPI & 1.00 & 1.76  & 5.87  & 18.75 & 22   \\ \addlinespace

\multirow{3}{*}{6} 
 & GB-MPC      & 0.00 & 11.29 & 4.75 & 11.75 & 80   \\
 & Vanilla MPPI     & 0.00 & 14.65 & -0.91 & 19.37 & 21   \\
 & COSMIK-MPPI & 0.00 & 11.99 & 5.28  & 69.57 & 21   \\ 
\bottomrule
\end{tabular}

  \vspace{-0.5cm}

\end{table}

Experiments were conducted on a workstation equipped with an Intel Core i9-14900K CPU and an NVIDIA RTX 5000 Ada GPU. Both GB-MPC and MPPI were evaluated using a prediction horizon of 50 steps at a control frequency of 50Hz. The GB-MPC implementation used 10 SQP iterations per control step, while MPPI was configured with 1000 rollouts.

Table~\ref{tab:predictive_collision_comparison} reports a comparative evaluation of GB-MPC, vanilla MPPI, and the proposed COSMIK-MPPI under increasing scenario complexity. Across Scenarios 1–3, all controllers achieve a 100\% success rate, indicating that the task remains tractable in lightly constrained environments. However, their computational behaviors differ significantly. GB-MPC solves Scenario 1 in 15ms, but its computation time increases by a factor of five in Scenarios 2 and 3, revealing sensitivity to added collision constraints. In contrast, both vanilla MPPI and COSMIK-MPPI maintain nearly constant computation times (21–22ms) across scenarios, highlighting a key advantage of sampling-based methods: their cost is largely independent of constraint complexity.

This improved scalability comes at the expense of reduced trajectory smoothness. At the planner level, MPPI’s stochastic sampling and importance-weighted updates introduce higher-frequency variations in the control sequence, whereas GB-MPC solves a structured OCP with quadratic regularization that naturally promotes smooth solutions. This difference is reinforced at the control level: GB-MPC computes time-varying Riccati gains and feedforward torques within the same optimization, ensuring planning–execution consistency, while MPPI relies on fixed PD gains with feedforward torques reconstructed via inverse dynamics, leading to higher torque variations as reported in Table~\ref{tab:predictive_collision_comparison}.

This divergence becomes critical in Scenarios 4 and 5, which feature non-convex geometries and multiple simultaneous avoidance constraints. In these cases, GB-MPC fails to compute feasible solutions within real-time limits, whereas both vanilla MPPI and COSMIK-MPPI successfully complete the tasks while maintaining constant computational performance. These results confirm that sampling-based methods scale more effectively than GB-MPC in the presence of dense and non-convex collision constraints.

Scenario 6 is intentionally infeasible and serves to evaluate controller behavior under persistent constraint violation. None of the methods reaches the target. GB-MPC and COSMIK-MPPI exhibit the most conservative behavior, prioritizing collision avoidance over goal convergence. 

Comparing the sampling-based methods, COSMIK-MPPI retains the same real-time efficiency as vanilla MPPI while improving robustness in constrained settings. It avoids the  collision violations observed with vanilla MPPI in Scenarios 6 and achieves comparable or better goal accuracy in most cases. By embedding safety directly through trajectory termination rather than large penalty weights, COSMIK-MPPI consistently biases the solution toward safer behaviors as complexity increases.

\section{Experimental validation}

\begin{figure*}[t]
  \centering
  \includegraphics[width=0.7\linewidth]{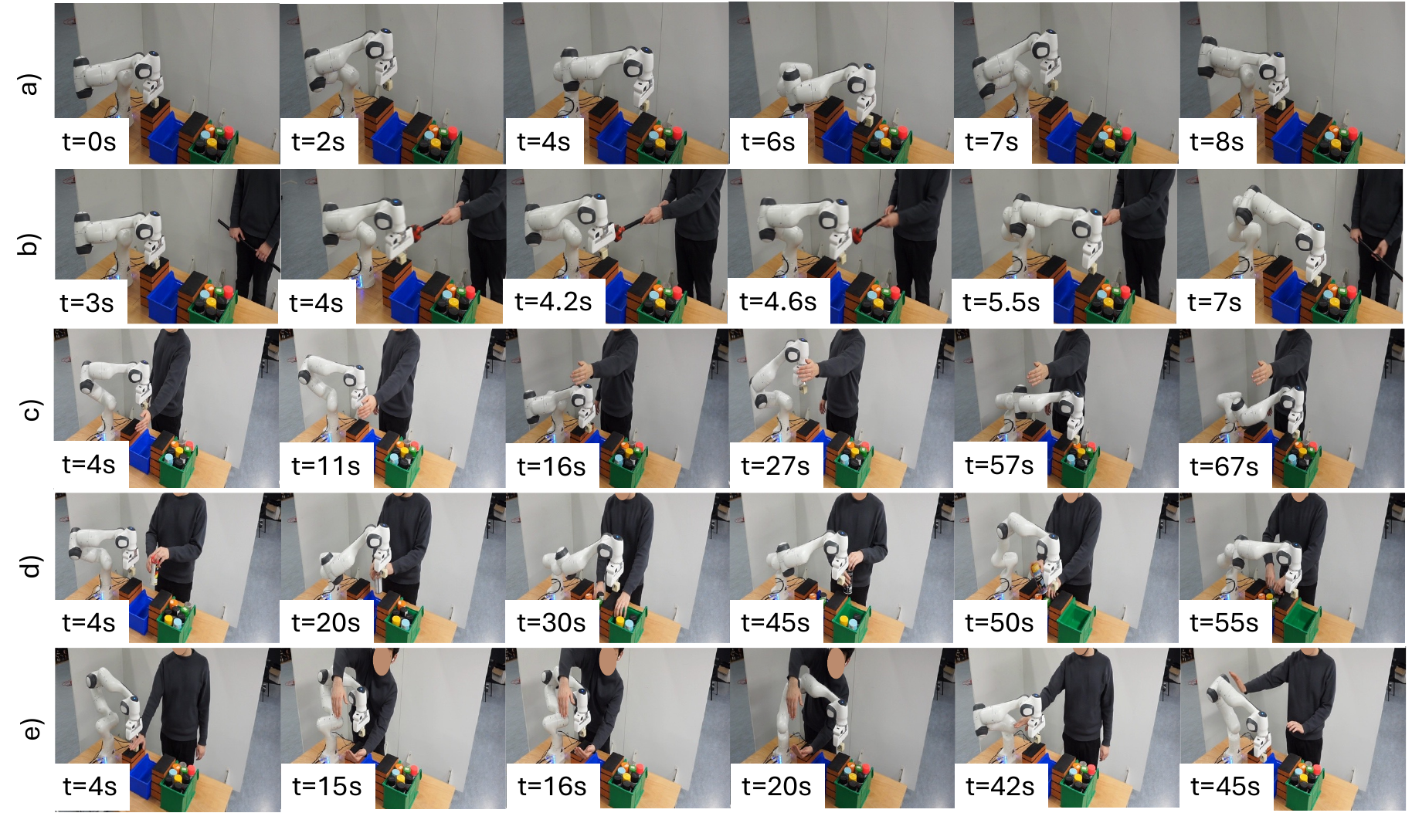}
  \caption{Five tasks used to assess COSMIK-MPPI safety and robustness for collision avoidance with with human. a) Static environment during baseline pick-and-place. b) Physical perturbations and compliance. c) Human-aware collision avoidance. d)  Concurrent human activity. e) Highly constrained human–robot interaction.  }
  \label{fig:exp}
  
  \vspace{-0.8cm}

\end{figure*}

To evaluate the proposed approach, we refer the reader to the accompanying video, which illustrates the system behavior in dynamic human–robot interaction in shared workspace scenarios. As summarized in Fig.~\ref{fig:exp} and further shown in the video, we consider a set of pick-and-place experiments of increasing complexity based on a common baseline task. The robot is instructed to grasp a cubic object placed in front of it and translate it by 30cm along the horizontal plane to a predefined target location. Five experimental scenarios are designed to progressively assess safety, robustness, and online replanning capabilities in human–robot interaction. All experiments share the same task parameters, initialization, number of collision pairs, and robot configuration, while the surrounding environment is modified through human interaction:

\begin{enumerate}[label=\alph*)]
\item Static environment:
A simple pick-and-place task performed in a static environment, serving as a baseline.
\item Physical perturbations and compliance:
During execution, the robot is physically perturbed using a rigid stick, evaluating its compliant behavior and ability to safely react to unexpected contacts while replanning to complete the task.
\item Human-aware collision avoidance with dynamic capsules:
A human operator deliberately blocks the robot motion with one arm. Human pose is estimated online using the RT-COSMIK framework to infer dynamic collision capsules. The robot autonomously replans and safely moves below the arm to complete the task.
\item Concurrent human activity:
The robot performs the pick-and-place task while a human simultaneously transfers spray paints between two bins in the same workspace. Human motion is unconstrained and does not explicitly account for the robot, introducing continuous and unpredictable workspace occupancy.
\item Highly constrained human–robot interaction:
The human forms a closed loop with both arms, creating a narrow passage and a non-convex collision setup. The robot must navigate through this constrained region to complete the task without contact.
\end{enumerate}

Results are reported for a single execution per scenario. As MPPI is stochastic and the experiments aim to probe distinct interaction modes rather than provide statistical validation, we focus on qualitative behavior, execution time, and motion characteristics. Reported times are indicative and used for relative comparison.

In the baseline static experiment (Fig.~\ref{fig:exp}(a)), the task was completed in 6s with a smooth, near-straight trajectory and no replanning, serving as a reference.

Under large physical perturbation (Fig.~\ref{fig:exp}(b)), execution time increased to 7.5s. The robot reacted compliantly, briefly slowing while continuing toward the goal along a diagonal path. The replanned motion remained smooth, without abrupt velocity or acceleration changes, demonstrating safe recovery through gradual adaptation.

In the concurrent human activity scenario (Fig.~\ref{fig:exp}(c)), execution time increased to 40s, mainly due to the time required for the human to complete the box-filling task and clear the target area, after which the robot finalized its motion. Frequent replanning led to back-and-forth adjustments as the robot adapted to the evolving workspace while maintaining safe operation.

 \begin{figure}[t]
  \centering
  \includegraphics[width=0.80\linewidth]{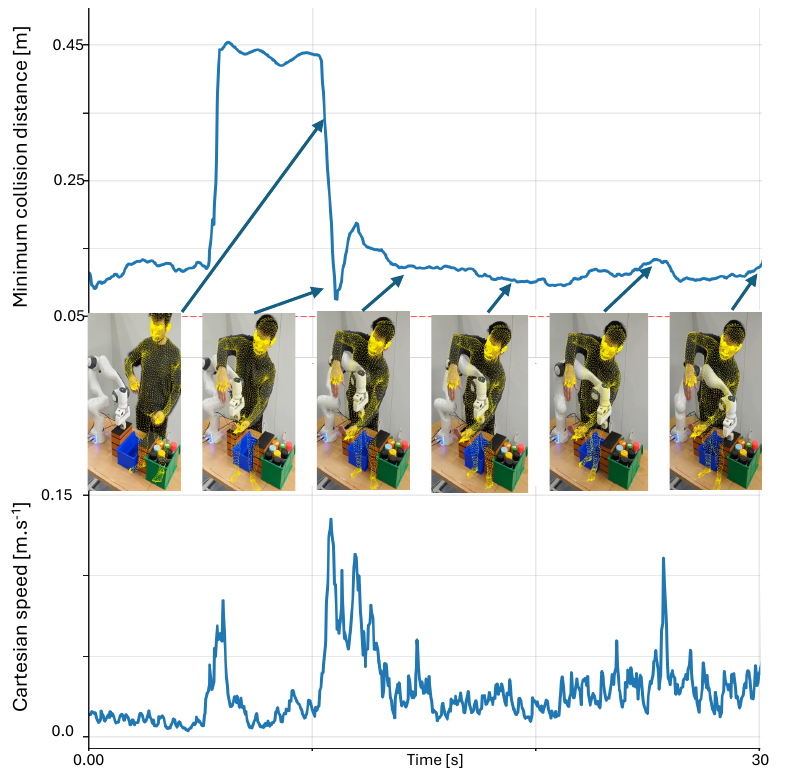}
 \caption{Analysis of the minimum collision distance between the robot and the environment and the human with the Cartesian end-effector velocity during scenario 5.}
  \label{fig:distance}
  
  \vspace{-0.5cm}

\end{figure}

In the human-aware collision avoidance scenario (Fig~\ref{fig:exp}.d), the execution time further increased to 50s. As shown in the accompanying video, the human actively blocked the robot motion before positioning the arm along the nominal trajectory. The robot successfully replanned online and passed below the arm. Despite close proximity, the motion remained smooth and safe. Near the end of execution, the robot approached a kinematic singularity, requiring additional reconfiguration; it again slowed down and generated a smooth, collision-free motion, deliberately trading speed for safety.

The final experiment, involving a highly constrained and non-convex collision scenario (Fig.~\ref{fig:exp}.e), in which the human formed a circle with both arms, was completed in 18s. As shown in Fig.~\ref{fig:distance}, the increased execution time was solely due to the robot deliberately reducing its speed while navigating the narrow passage. When the distance to the human became very small, the robot slowed down to reconfigure safely before accelerating again. As expected in this scenario, the minimum collision distance remained small throughout the task but collisions were never observed.

These experiments reveal a consistent qualitative behavior: as environment and interaction complexity increases, the robot naturally trades execution speed for safety. The proposed approach produces smooth and conservative motions, even in highly constrained situations, and leverages compliance and online replanning to ensure safe operation in close proximity to humans.

\section{Discussion}

The present paper benchmarked in simulation classical GB-MPC, vanilla MPPI, and the proposed COSMIK-MPPI for collision avoidance, and experimentally validate the use of COSMIK-MPPI for complex human robot shared workspace.

Simulation results demonstrate that while GB-MPC performs well in simple settings and often yields smoother motions, its performance rapidly degrades in the presence of multiple collision primitives and non-convex geometries. In such scenarios, GB-MPC fails to find feasible solutions within real-time constraints due to the growing number of active collision constraints. In contrast, MPPI-based methods maintain consistent real-time performance (21–22ms) and generate smooth, collision-free trajectories even in highly constrained environments. COSMIK-MPPI further improves robustness by embedding safety directly into the rollout termination, avoiding the need for large penalty weights while preserving motion smoothness and reactivity. These differences can be attributed to the underlying computational structure. The GB-MPC state-of-the-art implementation based on Crocoddyl~\cite{mastalli20crocoddyl} relies on constrained SQP optimization executed on a single CPU thread. As the number of collision constraints increases, the problem complexity grows rapidly, and although certain components can be parallelized, the core optimization loop remains sequential, limiting scalability. This issue has been previously reported~\cite{haffemayer2024model} and has motivated diffusion-based warm-start strategies for GB-MPC~\cite{Haffemayer2026DiffusionMPC,Oelerich2026SafeFlowMPC}. While GB-MPC provides attractive hard safety guarantees through explicit constraints, these guarantees come at the expense of scalability, particularly in dynamic and noisy human–robot interaction scenarios. In contrast, MPPI avoids warm-started optimization and accommodates numerous collision constraints through sampling and cost shaping. Although it lacks strict optimality guarantees, the experiments show that the trade-off is acceptable in practice as safety emerges from conservative motion generation, compliance, and continuous replanning. COSMIK-MPPI further improves robustness by terminating unsafe rollouts which result in more reliable behavior.

All the approaches rely on the robot’s intrinsic compliance, which is essential for safe human–robot workspace and is achieved by the low-level controller described in section~\ref{subsec:low_level_controller}. Compliance allows the robot to absorb physical perturbations (Experiment 2) and maintain smooth predictive planning without aggressive reactions.

Even if not the main topic of this study, an important feature of the proposed framework is the integration of RT-COSMIK \cite{rtcosmik2025} for markerless online human pose estimation, which directly supports the human–robot interaction scenarios in Experiments 3–5. RT-COSMIK achieves real-time performance and robustness to partial and dynamic occlusions through NLF, a property that is essential in close-proximity interactions involving self and robot-induced occlusions. The inverse kinematics stage used to infer dynamic collision capsules from the estimated pose further acts as a filtering layer, reducing noise and local inconsistencies in the collision geometry. While primarily leveraged here for collision avoidance, this perception pipeline also lays the groundwork for richer physical and task-level human–robot interaction. Additionally, our framework can be directly extended without additional development cost to take into account accurate predictions of human motion, which will improve further the efficiency of the interaction. 

This study demonstrates the safety and qualitative robustness of the proposed controller in controlled and semi-controlled scenarios. In future, extending it to a thorough assessment of robustness and efficiency ultimately requires large-scale evaluations with industrial operators or end users performing realistic tasks. The safe behavior exhibited by COSMIK-MPPI, even under close and highly constrained human–robot interaction, is a key enabler for such studies, as it allows extensive user evaluations without imposing restrictive assumptions on human behavior.

\bibliographystyle{IEEEtran}
\bibliography{references}

\end{document}